\documentclass[letterpaper, 10 pt, conference]{ieeeconf}  

\IEEEoverridecommandlockouts                              

\usepackage{array}
\usepackage{fancyhdr}
\usepackage{graphicx}
\usepackage{amsfonts}
\usepackage{amssymb}
\usepackage{amsmath}
\usepackage{verbatim}
\usepackage{color}

\usepackage{hyperref}
\hypersetup{
    pdffitwindow=false,     
    pdfstartview={FitH},    
    pdfnewwindow=true,      
    colorlinks=true,       
    linkcolor=red,          
    citecolor=green,        
    filecolor=magenta,      
    urlcolor=cyan           
}

\usepackage{cite}

\usepackage{listings}
\usepackage{hyperref} 
\usepackage[nolist,nohyperlinks]{acronym}

\usepackage{amsmath}
\usepackage{booktabs}
\usepackage{multicol}
\usepackage{multirow}

\usepackage[capitalise]{cleveref}

\usepackage{color} 
\usepackage{xcolor}
\hypersetup{
    colorlinks=true,
    linkcolor={red!70!green},  
    citecolor={blue!50!green}, 
    urlcolor={blue!80!black}   
}

\newcommand{\ie}{\textit{i}.\textit{e}., }

\usepackage[per-mode=symbol]{siunitx}

\usepackage{algorithm,algorithmic}
\newenvironment{conditions}
  {\par\vspace{\abovedisplayskip}\noindent\begin{tabular}{>{$}l<{$} @{${}={}$} l}}
  {\end{tabular}\par\vspace{\belowdisplayskip}}

\DeclareGraphicsExtensions{.pdf,.jpeg,.png}

\begin{document}
\frenchspacing

\title{\LARGE \bf Online Center of Mass Estimation for a \\Humanoid Wheeled Inverted Pendulum Robot}

\author{Munzir Zafar$^*$, Akash Patel$^*$, Bogdan Vlahov, Nathaniel Glaser, Sergio Aguilera, Seth Hutchinson
\thanks{$^{*}$Joint First Authors}%
\thanks{Munzir Zafar, Akash Patel, Bogdan Vlahov, Nathaniel Glaser, Sergio Aguilera and Seth Hutchinson are with the Institute of Robotics and Intelligent Machines at the Georgia Institute of Technology, Atlanta, GA, 30332, USA. email: {\tt\small mzafar7@gatech.edu, apatel435@gatech.edu, bvlahov3@gatech.edu, nglaser@gatech.edu, sfaguile@gatech.edu, seth@gatech.edu}}%
}

\begin{acronym}
\acro{CoM}{Center of Mass}
\acroplural{CoM}[CoMs]{Center of Masses}
\acro{WIP}{Wheeled Inverted Pendulum}
\acro{DoF}{Degree of Freedom}
\acroplural{DoF}[DoFs]{Degrees of Freedom}
\acro{RRT}{Randomly-Exploring Random Trees}
\acro{ESO}{Extended State Observer}
\acro{ADRC}{Active Disturbance Rejection Control}
\acro{PID}{Proportional-Integral-Derivative}
\acro{BLR}{Bayesian Linear Regression}
\acro{LQR}{Linear Quadratic Regulator}
\acro{DART}{Dynamic Animation and Robotics Toolkit}
\end{acronym}

\maketitle
\thispagestyle{empty}
\pagestyle{empty}
\begin{abstract}

We present a novel application of robust control and online learning for the balancing of a n \ac{DoF}, \ac{WIP} humanoid robot.
Our technique condenses the inaccuracies of a mass model into a \ac{CoM} error, balances despite this error, and uses online learning to update the mass model for a better \ac{CoM} estimate.
Using a simulated model of our robot, we meta-learn a set of excitory joint poses that makes our gradient descent algorithm quickly converge to an accurate \ac{CoM} estimate.
This simulated pipeline executes in a fully online fashion, using active disturbance rejection to address the mass errors that result from a steadily evolving mass model.
Experiments were performed on a 19 \ac{DoF} \ac{WIP}, in which we manually acquired the data for the learned set of poses and show that the mass model produced by a gradient descent produces a \ac{CoM} estimate that improves overall control and efficiency.
This work contributes to a greater corpus of whole body control on the Golem Krang humanoid robot.

\end{abstract}

\section{Introduction}
\label{sec:intro}
\acresetall{}

\begin{figure}[t]
	\centering
	\includegraphics[width=0.4\columnwidth, clip]{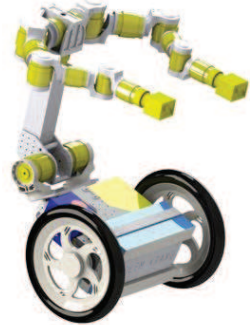}
	\vspace{-0.5\baselineskip}
   \caption{Full Body of our WIP Humanoid.}
   \label{fig: system1}
\end{figure}

Combining the maneuverability of a two-wheeled mobile platform and the dexterity of robotic arms, humanoid \ac{WIP} robots present novel challenges to the robotics research community.
Humanoid robot stabilization is fundamental to keep the robot safe and for the robot to accomplish higher-level objectives.
Furthermore, keeping a \ac{WIP}, such as the one presented in Fig.~\ref{fig: system1}, balanced is a fundamental task in which the controller needs to be constantly working and thus should be energy efficient \cite{Bature2014}.
Stabilization is usually accomplished through the control of a simplified two \ac{DoF} model which summarizes the \ac{CoM} of all the joints into one as shown in Fig~\ref{fig: system}. This simplification is usually done for both \ac{WIP} humanoid robots \cite{Zafar2016,canete2012disturbance, takei2009baggage}, as well as for legged humanoids \cite{Carpentier2016, Muscolo2011, Kudruss2015}.
All frameworks presented to stabilize \ac{WIP} robots consider that the mass and \ac{CoM} for each of the joints is accurately known \cite{SangJoo2007,Sihite2018,Pajon2017} to compute the simplified two \ac{DoF}  WIP model. However, the mass and real location of the \ac{CoM} is difficult to obtain, as robot systems can be complex and they might change throughout time. The discrepancy in the parameters of the robot affects the controller's performance, diminishing the robot's dexterity and increasing the power consumption.

Regarding these uncertainties in the model, one common control methodology uses the Modern Control Paradigm \cite{Gao2006} which focuses on the modeling of the system as, $\ddot{y} = f(y,\dot{y},w,t) + bu$, where $y$ is the position output and $w$ is an unknown input force. Once the system is modeled, it is approximated to a linear, time-invariant and disturbance-free model, to design a control law. This approach relies on the model approximation $\bar{f}(\dot{y},y)$ to be ``close enough'' to the real model in the neighborhood of the operation point. In \cite{Gao2006} and later in \cite{canete2012disturbance}, \acp{ESO} are used to estimate the modeled uncertainties and improve the control of the systems. The approach used collapses all the uncertainties and external forces under one element which is later eliminated through feedback control.  From an online learning approach, commonly used models rely on the knowledge and accuracy of the \ac{CoM} \cite{Luo2012,Chen2017,Yang2016}. Very few have worked on model parameter estimation such as \cite{Kim2016,Jamone2014}, but focus more on the estimation of external parameters such as terrain coefficient or external forces than on the robot itself. Finally, recent research involving mobile manipulators has focused on the use of \ac{ADRC} \cite{Jiang2016,Ruan2014,Wei2017} to control systems which use external uncertainties to conduct feedback control.

Our approach improves our model parameter estimation using the knowledge of the \ac{ESO} through online learning.
The goal of this framework is to create models that are improved upon by real-world systems and data. 
Given a model of our system, we want to improve the values of the parameters by measuring the disturbances of the system when it is not subject to external forces. Then, as the robot changes its joints position, we are able to update our parameters in an online fashion.
To accomplish this task, we propose the following methodology. Given an initial estimation of the parameters of our model $\beta_0$, we use \ac{ADRC} \cite{Gao2006} to estimate the error between the parameters estimated \ac{CoM} and the real one for different joint configurations. This error is used to update our knowledge of the model parameters through gradient descent. We show that this methodology works, but it might take numerous positions to converge. Thus, we propose a meta-learning algorithm to find the poses which induce the largest gradient step for gradient descent.
The main contributions of this works are: \textit{i)} a novel use of the \ac{ESO} and \ac{ADRC} to estimate the error in the model parameters; \textit{ii)} an online learning algorithm to update and improve our model parameters; \textit{iii)} a meta-learning framework to improve the speed and accuracy of our learning algorithm; \textit{iv)} and  preliminary results on a real robot with 19 \ac{DoF} that show the improvement of the system's performance.

This paper is organized as follows. Section \ref{sec:method} presents the \ac{WIP} robot and the methodology, as well as discusses the learning, meta-learning, and \ac{ADRC} techniques. Sections \ref{sec:experimental_design} and \ref{subsec:results} describe and present the different simulations and experiments. Finally, section \ref{sec:conclusion} presents the conclusions of our work.

\section{Methodology}
\label{sec:method}

\begin{figure}[htb]
	\centering
	\includegraphics[width=0.9\columnwidth]{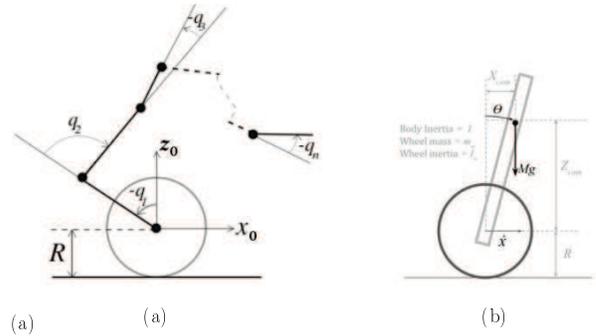}
    \vspace{-.5\baselineskip}
    \caption{Full Body of a typical WIP Humanoid with $n$ links (a) 2D Simplified Model (b)}
    \label{fig: system}
\end{figure}

The goal of the proposed approach is to improve the \ac{CoM} estimate of a \ac{WIP} Humanoid. A \ac{WIP} Humanoid is a highly redundant manipulator mounted on a differential wheeled drive able to dynamically balance itself in an inverted pendulum configuration (Fig \ref{fig: system}). A good estimate of the \ac{CoM} is important for any approach to control dynamically balancing humanoids. This is because the balancing task requires the \ac{CoM}'s ground projection to always lie in the support polygon. The support polygon of a WIP is a rectangle on width equal to the distance between the wheels and a small length given by the compression of the wheels against the ground. This support polygon is very thin, hence is important to decreasing the room for errors in \ac{CoM} estimates compared to, say, bipedal humanoids where support polygons are much larger.

Let us define frame $0$ as the frame where the origin  is located at the midpoint between the wheels with its $x$-axis always along the heading direction and $z$-axis always vertical. We are interested in the coordinates of the \ac{CoM} of the body in this frame. Specifically, we want the $x$-coordinate of body \ac{CoM} in this frame to be zero in order to balance the robot. Homogeneous coordinates of body CoM in frame $0$ are given by
\begin{align}
    X_{com}(q) \!= \!\begin{bmatrix} x_{com}(q) \\ y_{com}(q) \\ z_{com}(q) \\ 1 \end{bmatrix} &\!= \! \frac{\sum_{i}^{L} m_i X_i^0(q)}{\sum_{i}^{L} m_i} \!= \! \frac{\sum_{i}^{L} m_i {T}_{i}^{0}(q)X_i^i}{\sum_{i}^{L} m_i} \nonumber
\end{align}
where we are interested in the $x$ component of the CoM
\begin{align}
    x_{com}(q) &= \phi(q)^\top \beta \label{eq:xcom}
\end{align}
and the variables are described in Table~\ref{tab:Variables}.
\begin{table}[b]
    \centering
    \caption{\label{tab:Variables}System variables.}
    \begin{tabular}{c|p{0.79\columnwidth}}
    Variable & Description \\
    \hline
    $L$ & number of links in the body \\
    $q$ & $\begin{bmatrix} q_1 & ... & q_L \end{bmatrix}^\top$ position of all joints in the body \\
    $m_i$ & mass of link $i$ \\
    $X_i^0(q)$ & is CoM of link $i$ expressed in frame $0$ \\
    $X_i^i$ & $\begin{bmatrix} x_i & y_i & z_i & 1 \end{bmatrix}^\top$ local CoM of local frame $i$ \\
    $T_{i}^{0}(q)$ & transformation from frame $i$ to frame $0$ \\
    $\beta$ & $\begin{bmatrix} m_1X_1^{1\;\top} & ... & m_L X_L^{L\;\top} \end{bmatrix}^\top \in \mathcal{R}^{4L}$ \\
    $\phi(q)$ & $\begin{bmatrix} \phi_1(q) & ... & \phi_{4L}(q) \end{bmatrix}^\top$ feature vector of known geometric functions of $q$ \\
    \end{tabular}
\end{table}

$\beta$ is the set of unknown parameters comprising mass and mass times \ac{CoM} of individual links in the body. This choice of parameters is such that the parameters appear linearly in the model. Treating \ac{CoM} values independently from masses would make the model quadratic in parameters. Improving the estimate of the body's \ac{CoM} entails improving our knowledge of $\beta$. One way to achieve this is to disassemble the robot into individual links and perform physical measurements for mass and \ac{CoM} of each link. This is tedious and hence undesirable. However, the fact that the \ac{CoM} model is linear in $\beta$ allows us to use linear regression or gradient descent to improve our model parameter estimates. We choose gradient descent because of its ability to enforce physical consistency constraints. For one, it converges to $\beta$ values in the neighborhood of the initial $\beta$ which are more likely to be physically consistent as opposed to linear regression which might learn solutions that fit the data but are physically nonsensical. Secondly, constraints such as total body mass can be explicitly enforced in the learning process through the use of Lagrange multipliers. The details of this appear in \cref{subsec: learning}.

These learning techniques rely on our ability to collect data for poses $q$ and corresponding values of outputs $x_{com}(q)$. The simplest way to collect this data is to make use of the fact that in the ideal case, $x_{com}(q) = 0$ when the robot is in a balanced state. Assuming that all joints in the body  shown in Fig. 1b can be locked at a specific pose $\{$ $q_2$, $...$, $q_L$ $\}$, there exists a position for $q_1$ (the base link) that can balance the robot. We can collect data offline by manually moving $q_1$ such that the robot is in a balanced state. However, this is again tedious; performing the same job online would avoid this labor. To this end, we utilize \ac{ADRC} \cite{canete2012disturbance} to balance the robot despite a bad estimate of body \ac{CoM}, the details of which appear in \cref{subsec: online}. One may ask: Why the need to improve  the \ac{CoM} model if there already exists a controller that is able to stabilize the robot despite a bad \ac{CoM} estimate? The answer to this is twofold: Firstly, \ac{ADRC} achieves balancing but is inefficient, i.e. it takes more time and aggressive control inputs to stabilize a bad estimate of \ac{CoM}. Secondly, \ac{ADRC} works only when controlling a single rigid link on wheels which is the case when body joints are locked. If however the joints are unlocked, more complex controllers are needed that rely on an accurate estimate of the \ac{CoM}.

We have so far discussed how to obtain the value of $x_{com}(q)$ at any give pose $q$. It is important to determine what poses at which we should collect this data. This is because with a highly redundant system, the configuration space is too large and relying on arbitrary poses may make the learning process inefficient and time-consuming. We choose poses such that every next pose causes the largest average gradient descent step over a large set of randomly chosen erroneous $\beta$ estimates. This is discussed in \cref{subsec: learning}.


\subsection{Learning Algorithm} \label{subsec: learning}

For the learning algorithm, we make use of gradient descent. The objective function to be minimized is determined based on the fact the $x$-component of CoM should be zero in a balanced pose. In order to make the cost function locally convex with respect to $\beta$, we aim to minimize the square of the $x$-component of CoM.
\begin{align}
    J(\beta) &= \tfrac{1}{2} \left[ x_{com}(q; \beta) \right]^2 \nonumber \\
    &= \tfrac{1}{2} \beta^\top \phi(q) \phi(q)^\top \beta
\end{align}
where we have made use of the definition of $x_{com}$ in \eqref{eq:xcom}.

The gradient with respect to $\beta$ will therefore be
\begin{equation}
    \nabla_\beta J(\beta) = \phi(q) \phi(q)^\top \beta
\end{equation}
The update step will be
\begin{equation}
    \beta_{t+1} \leftarrow \beta_t - \eta \nabla_\beta J(\beta_t) \label{eq:update}
\end{equation}
where $\eta$ is the step-size, which is a hand-tuned parameter. We begin with an initial estimate of $\beta$. As data for the new balanced pose $q$ is collected, we make use of the gradient update step in \eqref{eq:update} to improve $\beta$ estimates. This is repeated until $\phi(q)^\top \beta$ consistently drops below a threshold $x_{tol}$ for a few iterations.

\subsection{Meta-Learning Algorithm} \label{subsec: meta-learning}

We also deal with the problem of determining a training set of poses that makes the learning process efficient or less time-consuming. For robots with many \aclp{DoF}, the configuration space is huge and choosing an arbitrary set of training poses will likely make the learning inefficient. We determine this training set offline, only using the model in simulation, using the algorithm presented in \cref{algo: poses}. The algorithm requires a large pool of randomly  generated balanced and safe poses $\bar{q} \in \mathcal{R}^{n_{DOF} \times n_{poses}}$. A balanced pose is one where a ``real'' robot (\ie with $\beta$ values we pretend to be real) is balanced. A safe pose is one where the robot does not collide with itself or the ground, and the joint values are within their physical limits. We precompute the numerical values of the feature vector $\phi(q)$ evaluated at each pose in $\bar{q}$ and store them in $\Phi \in \mathcal{R}^{dim(\beta) \times n_{poses}}$. The algorithm also requires a set of randomly generated erroneous $\beta$ vectors: $\bar\beta \in \mathcal{R}^{dim(\beta) \times n_{\beta}}$. This is done by choosing values of $\beta$ vectors that cause $x_{com}$ estimate errors in estimating the ``real'' robot's \ac{CoM} to be of the same order as is observed in the physical system.
\begin{algorithm}
    \caption{Pose Filtering}
    \begin{algorithmic}[1]
        \renewcommand{\algorithmicrequire}{\textbf{Input:}}
        \renewcommand{\algorithmicensure}{\textbf{Output:}}
        \REQUIRE Set of randomly generated safe \& balanced poses: $\bar{q} \in
        \mathbb{R}^{n_{DOF} \times n_{poses}}$,
        \newline Set of $\phi(q)$ evaluated at each given pose: $\Phi \in
        \mathbb{R}^{dim(\beta) \times n_{poses}}$,
        \newline Set of randomly generated erroneous $\beta$s: $\bar{\beta} \in \mathbb{R}^{dim(\beta)
        \times n_{\beta}}$
        \ENSURE Filtered set of poses: $\widetilde{q}$
        \REPEAT
        \STATE $i^* \leftarrow \underset{i \in \{1, ..., n_{poses}\}}{\mathrm{argmax}}
        \sum_{k}^{n_{\beta}} | \Phi_i^\top \beta_k |$
        \STATE $\widetilde{q} \leftarrow [\begin{matrix} \widetilde{q} &
        \bar{q}_{i^*} \end{matrix}] $
        \STATE $\phi^* \leftarrow \Phi_{i^{*}}$
        \STATE $\beta_k \leftarrow \beta_k - \eta \; \phi^* \phi^{*\top} \beta_k \quad \forall \quad k \in \{1, ..., n_\beta \}$
        \STATE $\Phi \leftarrow  \Phi \setminus \Phi_{i^*} $
        \STATE $\bar{q} \leftarrow \bar{q} \setminus \bar{q}_{i^*} $
        \UNTIL {$ | \phi^{*\top} \beta_k | < x_{tol} \quad \forall \quad k \in \{1,
        ..., n_\beta \} $ for last few iterations}
        \RETURN $\widetilde{q}$
    \end{algorithmic}
    \label{algo: poses}
\end{algorithm}
The key step in Algorithm \ref{algo: poses} is step 2 where the pose that causes the largest average error on all erroneous $\beta$'s is chosen to be added to the filtered set of poses $\widetilde{q}$ which is the output of the algorithm. This pose is also used to perform gradient descent on all $\beta \in \bar{\beta}$ (step 5). We choose the pose that causes the largest prediction error over the updated set $\bar\beta$ in each iteration because it is the most informative for the learning process. The learning process stops when the prediction errors due to all $\beta \in \bar{\beta}$ consistently fall below some tolerance $x_{tol}$ for a set number of iterations.



Even though the set of poses generated from meta-learning were acquired from different $\beta$s than that of the real robot, these poses generated a large error that then helped our entire $\bar{\beta}$ set to converge. If our robot's initial $\beta$ is in or even close to the set $\bar{\beta}$, these poses should have a similar effect and cause it to converge.




\subsection{Online Data Collection} \label{subsec: online}
We now discuss the problem of balancing the robot despite a bad estimate of body \ac{CoM} to obtain data points for the learning process. Given that body joints are locked at the desired pose $\{$ $q_2$, $...$, $q_L$ $\}$, the robot is equivalent to a single rigid link on two wheels, to be balanced by manipulating the base link $q_1$ and the wheels. We utilize \ac{ADRC} \cite{canete2012disturbance} for this purpose. This approach for balancing control of a \ac{WIP} Humanoid is originally intended to handle disturbances represented by a torque $\tau_D$ about the wheel axis. To see how this approach is applicable for our case, we can imagine a virtual robot that has $\beta$ values equal to our current bad estimate and is experiencing a disturbance torque such that the effective \ac{CoM} of the virtual system has shifted to the real \ac{CoM} of the physical system. Thus the problem of controlling a robot with a bad \ac{CoM} estimate is equivalent to one experiencing a disturbance torque about its wheel axle.

A brief explanation of the technique as it applies to our system is as follows. Linearizing the dynamics of WIP Humanoid with its joints locked at pose $q$ in a 2 \ac{DoF} system
\begin{align}
    \dot{X} = \frac{d}{dt} \begin{bmatrix} x & \dot{x} & \theta & \dot\theta \end{bmatrix}^\top = A(q)X + B(q)\tau_w
\end{align}
where
\begin{conditions}
x, \dot{x} & position and heading speed of the robot \\
\theta, \dot\theta & ang. position and speed of CoM about wheel axis \\
\tau_w & sum of torques applied on both wheels \\
B(q) & $\begin{bmatrix} 0 & 0 & b_x(q) & b_{\theta}(q) \end{bmatrix}^\top$
\end{conditions}
Note that $A$, $b_x$ and $b_{\theta}$ are functions of parameters such as \ac{CoM} distance from wheel axis and body inertia that are dependent on $q$. Applying LQR on this pose-dependent linearized system $(A(q), B(q))$ results in pose-dependent feedback gains
\begin{align}
    F(q) &= \begin{bmatrix} F_x(q)^\top & F_\theta(q)^\top \end{bmatrix}^\top = \mbox{LQR}(A(q), B(q))
\end{align}
Treating $\dot{x}$ and $\dot{\theta}$ dynamics as two independent subsystems by following \cite{miklosovic2005dynamic}, we can find the control inputs as
\begin{align}
    u_x = -F_x(q)^\top \begin{bmatrix} x & \dot{x} \end{bmatrix}^\top \quad u_\theta = -F_\theta(q)^\top \begin{bmatrix} \theta & \dot{\theta} \end{bmatrix}^\top\nonumber
\end{align}
The standard feedback control setting for \ac{WIP} systems has the control input defined by $\tau_w = u_x + u_{\theta}$. However, the key to perform active disturbance rejection is to estimate the numerical value of dynamic disturbances in the two subsystems,~$\hat{f_x}$~and~$\hat{f_{\theta}}$, due to the inaccurate \ac{CoM} estimate and compensate for those disturbances using feedback linearization:
\begin{align}
    \tau_w &= \left( u_x - \frac{\hat{f}_x}{b_x(q)} \right) + \left( u_\theta - \frac{\hat{f}_\theta}{b_\theta(q)} \right)
\end{align}
Here, $\hat{f_x}$ and $\hat{f_{\theta}}$ are estimating the dynamic disturbances $f_x$ and $f_{\theta}$ in the subsystems appearing in state space representation of the dynamic model
\begin{align}
    \ddot{x} &= f_x(X, q, \tau_D, u_\theta) + b_x(q)u_x \nonumber \\
    \ddot{\theta} &= f_\theta(X, q, \tau_D, u_x) + b_\theta(q)u_\theta
\end{align}
The estimates are found using \aclp{ESO}
\small{
\begin{align}
    \frac{d}{dt} \!\! \begin{bmatrix} \hat\theta \\ \hat{\dot{\theta}} \\ \hat{f}_\theta \end{bmatrix} &\!\!= \!\! \begin{bmatrix} \hat{\dot{\theta}} + l_{\theta 1}(\theta-\hat\theta) \\ \hat{f}_\theta + l_{\theta 2}(\theta - \hat\theta) \\ l_{\theta 3} (\theta - \hat\theta) \end{bmatrix}\!\!, &\!\!\!\!\!
    \frac{d}{dt} \!\! \begin{bmatrix} \hat{x} \\ \hat{\dot{x}} \\ \hat{f}_x \end{bmatrix} &\!\!=\!\! \begin{bmatrix} \hat{\dot{x}} + l_{x1}(x-\hat{x}) \\ \hat{f}_x + l_{x2}(x - \hat{x}) \\ l_{x3} (x - \hat{x}) \end{bmatrix}
\end{align}}
\normalsize
where the observer gains $l_x$ and $l_{\theta}$ are designed using pole placement.
\section{Simulation Results}
\label{sec:experimental_design}



We started experiments by simulating our pipeline; we first considered a \ac{WIP} model with 7 DoF in Matlab and next a \ac{WIP} with 19 DoF in the 3D \ac{DART} \cite{DART}.  The former served as a more tractable proof of concept that led into the latter, a more faithful representation of the robot that we will be using during the experiments.  In both simulations, we provided the class methods for instantiating an $L$-link \ac{WIP} model, updating their mass parameters $\beta$, approximating their dynamics, applying control, and visualizing the results. Simulation provided us with two key benefits over hardware: (1) it allowed us to rapidly spawn, control, and respawn our robot in a safe, realistic setting; and (2) it allowed us immediate access to parameters that were otherwise ``unknowable'', or difficult to obtain.  For our system specifically, these parameters are the masses and \aclp{CoM} for individual links, which are both numerous and inaccessible to measurements.
\begin{figure*}[!h]
\centering
\includegraphics[width=0.9\linewidth]{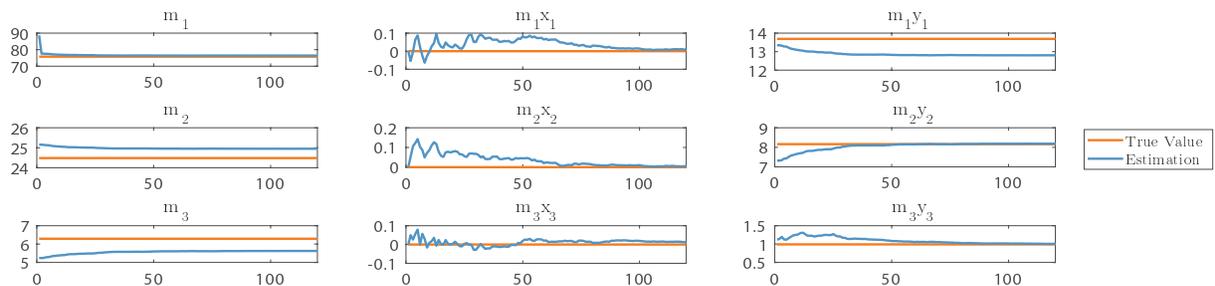}
\vspace{-.5\baselineskip}
\caption{\label{fig:Values} Values of the different parameters of the estimated model over 120 different configurations. The red lines show the real value of the parameter for the real robot, while the blue lines show the learned weights.}
\end{figure*}
To evaluate the performance of our algorithms we instantiated two full $L$-link \ac{WIP} models -- a ground truth model and an inaccurate model with an estimation of the parameters of the first robot. These two models served as placeholders for the arm's configuration and mass parameters. We then simplified these two models into their single link representation (Fig.~\ref{fig: system} - right). In Matlab, using an ODE45 integration loop, we simulated the system dynamics from the ground truth model and then calculated the control signals based on the estimated simplified model. In the \ac{DART} implementation, the dynamics were updated automatically by the simulator. We first started by tuning our \ac{ADRC}'s LQR gains to be able to control the estimated simplified model to the balance position of the ground truth model. During this process, we iteratively set both models to randomized joint angles on the configuration space. After tuning the controller and observer parameters for each joints configuration, the \ac{ADRC} would balance the systems to its true balance position, i.e. for a given configuration $q_2, q_3, \dots, q_L$, the \ac{ADRC} would find the value of $q_1$ that balanced the system.

\subsection{Gradient Descent Simulation}
\label{subsec:sim_results}

The offset given by the \ac{ADRC} for the estimated model was used in a gradient descent algorithm to update our estimated model parameters. Starting with the Matlab simulation, the estimated model was subject to initial noise for the initial estimation of $20\%$ from the real values of the parameters $m_i$, $m_ix_i$ and $m_iy_i$. Since each link had different properties (similar to our experimental robot), the noise perturbation differed; the first link has an approximated mass of $70 kg$  which gives a noise around $14kg$, while the third link has a mass of $6kg$ which give us a noise around $1.2kg$. Using \cref{eq:gradient_Descent} we update our $\beta$ for each iteration. A subset of the parameters of $\beta$ are shown in \cref{fig:Values}.


It can be seen in \cref{fig:Values} that our algorithm modifies the $\beta$ vectors, reaching a local minimum. For some parameters (as $m_1$, $m_1y_2$ or $m_3y_3$) the estimated values converge to the real values, while for others (as $m_2$, $m_3$ or $m_1y_1$) the values converge to a constant error. Even though we are finding a local minima and not necessarily the correct values, we will show that our new estimate of $\beta$ improves upon the initial values. After running different simulations, we notice that while the system always reaches a $x_{CoM}$ error of zero, the weights converge to different values -- giving the intuition that the system consists of several local minima.

This method has shown that the approach works in finding a better set of values than the ones we initially started with, but might not get to the global optimum (the real values). We think that this happens because of the nonlinearities of the system and because the $\beta$ vector is not perfectly decoupled to the value of the masses.

\subsection{Meta-learning for Gradient Descent Convergence}
\label{subsec:Meta-learning}
\begin{figure} [htb]
\centering
\includegraphics[width=0.8\columnwidth]{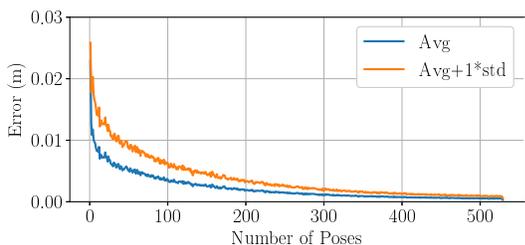}
\vspace{-0.8\baselineskip}
\caption{\label{fig:Error_sim} Mean Error of several $\beta$s through the learning algorithm for the meta-learned best 500 poses.}
\end{figure}

As described in section \ref{subsec: meta-learning}, we simulated 20,000 poses over 500 erroneous $\beta$s, and got a set of 528 poses until the error was $2mm$. Without using the meta-learning algorithm this process takes  over 5,000 poses. The result of our simulated learning curve is presented in \cref{fig:Error_sim}. We tested for several initial erroneous $\beta$s which started with an $x_{CoM}$ error of at least $2 cm$ with a standard deviation of $0.5 cm$. It can be seen that after $500$ updates using the optimal poses, the mean error decreased to almost $0 cm$, specifically the max $\beta$ error decreased to $x_{tol} = 2mm$.

\section{Experimental Results}
\label{subsec:results}

\begin{figure}
\centering
\includegraphics[width=0.9\columnwidth]{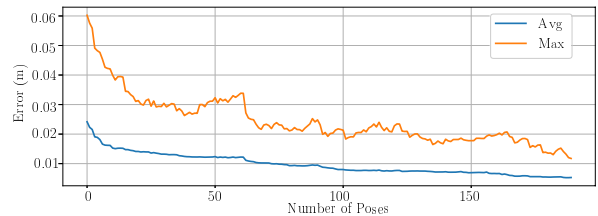}
\vspace{-0.2cm}
\caption{\label{fig:Error_hardware} Error in the parameters as we update the weights $\beta$ for different random configurations.}
\end{figure}

For the robot that we are using, Golem Krang \cite{Stilman2010}, determining its mass model link-by-link is intractable.  Furthermore, the summarizing \ac{CoM} described in \ref{sec:method} is difficult to obtain.  Instead of extracting the full mass model or \ac{CoM} estimates, we follow the procedure of other work \cite{bature2014comparison,khosla2013performance} to evaluate balancing performance.  Where the authors analyze more readily observable phenomena, such as distance traveled, time spent stabilizing, and power consumption.  In our case, these quantities were used to analyze whether or not subsequent refinements of an initial offset estimation $(\beta_0)$ improves the stabilizing control.
The physical experiments were separated in two parts: manual data collection and controller efficiency testing.
For the first part, we collected data from our subset of pre-determined balanced poses -- we manually positioned the robot in the first 236 poses acquired from the meta-learning algorithm in \cref{subsec: meta-learning} and calculated the error between the real $x_{com}$ and the estimation.  We obtained this error by setting the robot to presumed balanced pose (which may not actually be balanced under our inaccurate $\beta_0$), and adjusted the base link angle $q_1$ until the system became balanced.  We then separated this data into a training set of 190 poses and a testing set of 46 poses.  Then, using the training set, we implemented gradient descent to obtain a series of betas going from $\beta_1, \beta_2, \dots, \beta_{190}$. For each beta, we computed the errors produced by the remaining balanced poses in the testing dataset; the results are shown in Fig.~\ref{fig:Error_hardware}.
For $\beta_0$, we started with a mean error of $2.5 cm$ in the $x_{CoM}$ for the given $46$ poses and a maximum error of $6 cm$. With subsequent iterations, the mean error and the maximum error decreased. For $\beta_{190}$ we achieved a mean error of $0.4 cm$ with a maximum error of $1.2 cm$ for any given pose in the testing set.

For the second part, we used five of our learned $\beta$s to balance the robot in a given pose.  Specifically, we looked at the initial balancing action, which involves transitioning between a stable sitting position to an inverted pendulum position. For this action, the robot stands from three points of contact with the ground (two active wheels and a caster wheel). Then it rotates its wheels (at a speed which depends on its \ac{CoM} estimate) to lift off the caster, and it finally balances as a two-wheeled \ac{WIP}.
The balancing experiments tested different $\beta$ estimates to show how the overall control improves during the transition and steady state of the robot.
To investigate the connection between updated $\beta$ vectors and controller performance, we show the results of testing $\beta_{16}$, $\beta_{32}$, $\beta_{64}$, $\beta_{128}$ and $\beta_{190}$.  Smaller $\beta$s are not shown, since the robot controller was not able to securely stabilize the system.  Additionally, for each $\beta_i$ we tested seven attempts to see the reproducibility of the results.

\begin{figure}[h]
\centering
\includegraphics[width = 0.99\columnwidth]{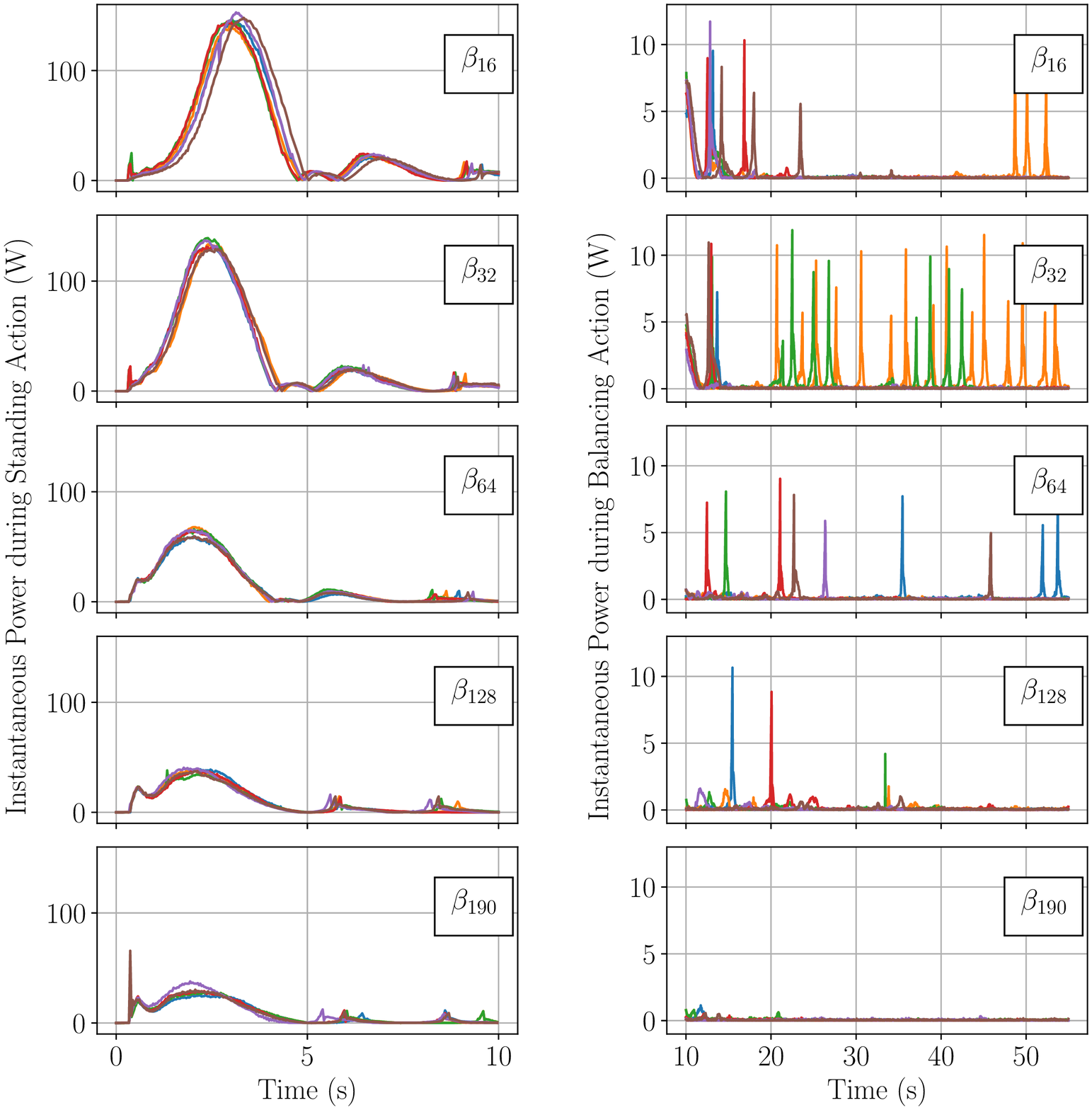}
\vspace{-1.5\baselineskip}
\caption{\label{fig:betas_comparison} Instantaneous power applied by the wheel motors.  Each plot includes the results corresponding to 7 independent runs for different values of $\beta$ ($\beta_{16}$, $\beta_{32}$, $\beta_{64}$, $\beta_{128}$ and $\beta_{190}$). The left column summarizes the sitting-standing transition (the first 10 seconds of the experiment), and the right column summarizes the \ac{WIP} balancing (the subsequent 10 to 60 seconds).}
\end{figure}

The instantaneous power consumption of the wheel motors during and after the transition to standing is shown in \cref{fig:betas_comparison}, and a summary of the control performance is presented in Table~\ref{tab:control_performance}.  The instantaneous power was calculated by multiplying the torque and angular velocity of the wheels.

\begin{table}[h!]
\centering
\caption{\label{tab:control_performance} Summary of the control performance under different betas.}
\vspace{-\baselineskip}
\begin{tabular}{c|p{0.9cm}|p{1.0cm}|p{1.4cm}|p{1.0cm}|p{1.5cm}}
$\beta$ & Max Pos. [m] & Resting Pos. [m] & Time until Resting [s] & Max Power [W] & Avg. Resting Power [mW] \\
\hline
\hline
$\beta_{16}$ &  $4.70 \pm 0.16$ & $2.49 \pm 0.03$ & $11.5 \pm 1.5$  &  $145 \pm 4$ & $7.83 \pm 1.69$\\
\hline
$\beta_{32}$ & $4.59 \pm 0.11$ & $2.67 \pm 0.11$ & $10.1 \pm 1.1$  &  $133 \pm 4$ & $8.85 \pm 7.09$\\
\hline
$\beta_{64}$ & $3.59 \pm 0.17$ & $1.53 \pm 0.05$ & $7.59 \pm 1.33$  &  $63.1 \pm 3.3$ & $2.95 \pm 1.04$\\
\hline
$\beta_{128}$ & $2.74 \pm 0.07$ & $1.13 \pm 0.03$ & $6.80 \pm 1.20$  &  $41.3 \pm 6.9$ & $2.90 \pm 0.60$\\
\hline
$\beta_{190}$ & $\bf 2.61 \pm 0.08$ & $\bf 1.08 \pm 0.03$ & $\bf 7.09 \pm 1.97$  & $\bf 34.5 \pm 13.2$ & $\bf 1.54 \pm 0.25$ \\
\end{tabular}
\end{table}


As shown in the left column of Fig.~\ref{fig:betas_comparison} and in Table~\ref{tab:control_performance}, the peak power consumption decreases with subsequent values of beta.  As shown in the right column of the same figure, the number of balancing adjustments (spikes in power consumption) is similarly reduced.  For the first $\beta$ values, the system occasionally destabilized and readjusted, whereas the latest $\beta_{190}$ value kept these adjustments and hence overall power consumption to a minimum.

Table~\ref{tab:control_performance} shows improvement in several quantities that characterize control performance: the initial overshoot position decreases by $44\%$ between the $\beta_{16}$ and $\beta_{190}$ iterations; the resting position decreases by $57\%$; the time until resting decreases by $38\%$; the peak instantaneous power decreases by $76\%$; and the average power during steady state balancing decreases by $80\%$.  Each of our performance metrics improves with more refined mass model parameters.  Together, these trends support the claim that the \ac{CoM} estimation procedure does improve balancing for a \ac{WIP}.


\section{Conclusion}
\label{sec:conclusion}
We have shown that the proposed methodology improves the \ac{CoM} estimate of a \ac{WIP} Humanoid and that these improvements translate to improved controller performance.
In simulation, using active disturbance rejection control, our robot successfully balances with an inaccurate prior mass model, collects new pose data at balanced positions, and learns from these poses to produce a more accurate \ac{CoM} estimate.
In hardware, we demonstrate that these refined estimates directly translate into improved controller performance.
Together, our simulation and hardware results support the claim that our algorithm -- a semi-automated, tractable procedure that refines the latent space mass model of a high dimensional system with few physically observable parameters -- does improve overall balance.  The algorithm was probed in simulation and verified physically on a 19 \ac{DoF} \ac{WIP} robot.
Our future work will implement the fully automated estimation pipeline--active disturbance rejection control, balanced pose data collection, and online learning--in an entirely online fashion on the physical robot, where it will improve its parameter estimates through meta-learned poses.




\bibliographystyle{IEEEtran}
\bibliography{IEEEabrv,onlinecom}
\end{document}